\pdfoutput=1

\documentclass[11pt]{article}

\usepackage[preprint]{acl}

\usepackage{times}
\usepackage{latexsym}

\usepackage[T1]{fontenc}

\usepackage[utf8]{inputenc}

\usepackage{microtype}

\usepackage{inconsolata}

\usepackage{graphicx}

\usepackage{booktabs}
\usepackage{hyperref}
\usepackage{amsfonts}
\usepackage{amssymb}
\usepackage{amsmath}
\usepackage{colortbl}
\usepackage{xcolor}
\usepackage[normalem]{ulem}
\usepackage{pifont}

\title{Multi-Hop Question Generation via Dual-Perspective Keyword Guidance}

\author{
 \textbf{Maodong Li\textsuperscript{1}},
 \textbf{Longyin Zhang\textsuperscript{2}},
 \textbf{Fang Kong\textsuperscript{1}}\thanks{Corresponding author}
\\
 \textsuperscript{1}School of Computer Science and Technology, Soochow University, China \\
 \textsuperscript{2}Institute for Infocomm Research, A*STAR, Singapore
\\
 \texttt{\{20234227022@stu,kongfang@\}suda.edu.cn} \quad
 \texttt{zhang\_longyin@i2r.a-star.edu.sg}
}

\begin{document}
\maketitle

\begin{abstract}
Multi-hop question generation (MQG) aims to generate questions that require synthesizing multiple information snippets from documents to derive target answers. The primary challenge lies in effectively pinpointing crucial information snippets related to question-answer (QA) pairs, typically relying on keywords. However, existing works fail to fully utilize the guiding potential of keywords and neglect to differentiate the distinct roles of question-specific and document-specific keywords. To address this, we define dual-perspective keywords—question and document keywords—and propose a \textbf{D}ual-\textbf{P}erspective \textbf{K}eyword-\textbf{G}uided (DPKG) framework, which seamlessly integrates keywords into the multi-hop question generation process. We argue that question keywords capture the questioner's intent, whereas document keywords reflect the content related to the QA pair. Functionally, question and document keywords work together to pinpoint essential information snippets in the document, with question keywords required to appear in the generated question. The DPKG framework consists of an expanded transformer encoder and two answer-aware transformer decoders for keyword and question generation, respectively. Extensive experiments demonstrate the effectiveness of our work, showcasing its promising performance and underscoring its significant value in the MQG task.
\end{abstract}
\section{Introduction}
Question generation (QG) is the task of generating questions based on given documents and target answers \cite{10.1145/3468889,ushio-etal-2023-practical}. It is a vital area in natural language processing (NLP), with considerable potential for applications in automated question answering, reading comprehension, and related fields \cite{liang-etal-2023-prompting,ijcai2024p889}. Compared to the QG task, multi-hop question generation (MQG) aims to generate questions that require synthesizing multiple information snippets from documents to derive target answers, making it more representative of real-world scenarios \cite{fei-etal-2021-iterative,fei-etal-2022-cqg,chen2023toward}.

\begin{figure}[t]
  \includegraphics[width=\columnwidth]{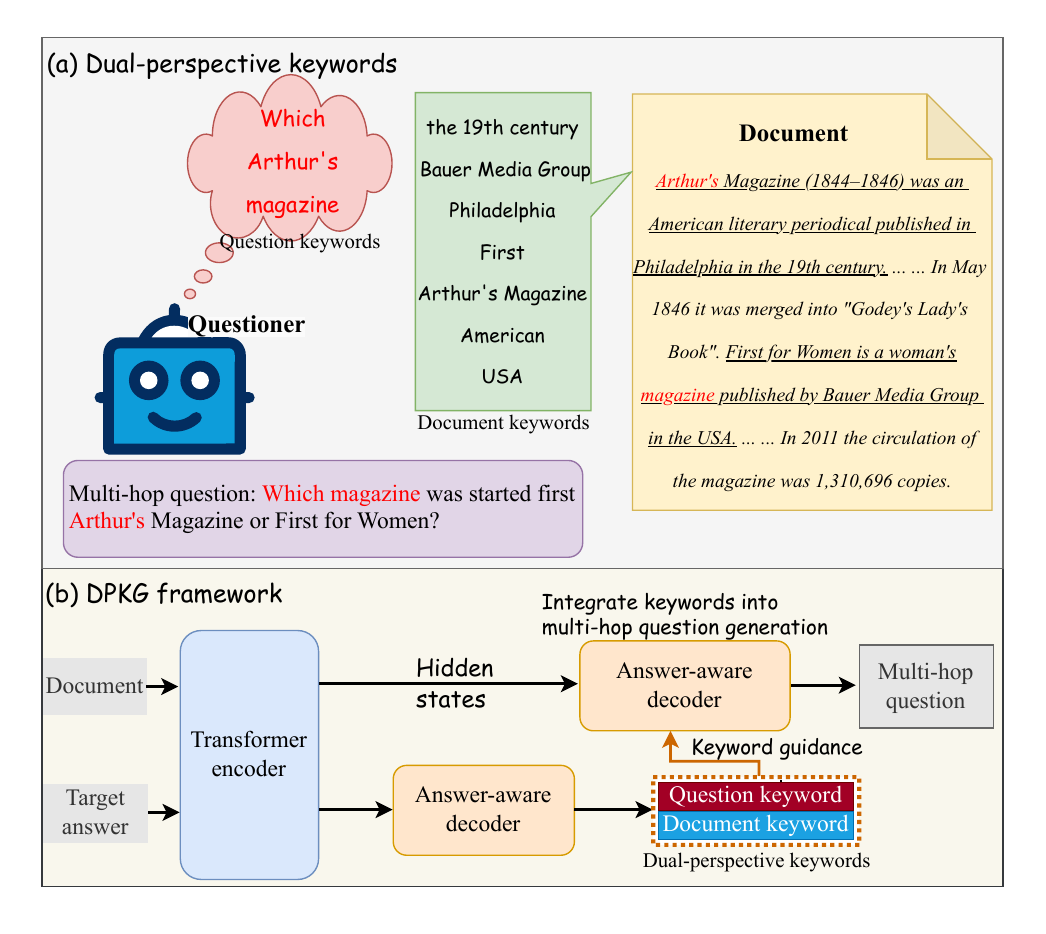}
  \caption{(a) shows the definition of dual-perspective keywords: question keywords and document keywords. The \textcolor{red}{red} highlights the question keywords, while the \underline{underlined text} indicates crucial information snippets related to question-answer pairs. (b) illustrates the proposed DPKG framework.}
  \label{fig01}
\end{figure}

The primary challenge in MQG is effectively pinpointing crucial information snippets related to question-answer (QA) pairs in documents, a task that many studies have approached using keywords \cite{DBLP:conf/emnlp/XiaG0YHLN23,ding-etal-2024-sgcm,liu2024coferagcomprehensivefullchainevaluation,fazili2024gensco}. However, current research does not fully leverage the guiding potential of keywords. Some studies focus exclusively on document-specific keywords \cite{su-etal-2020-multi,cao-wang-2021-controllable}, while others, including more recent, higher-performing models \cite{fei-etal-2022-cqg,fei-etal-2022-lfkqg} consider constraints on question-specific keywords during question generation. According to our understanding, as illustrated in Figure \ref{fig01}(a), question keywords reflect the questioner's intent, whereas document keywords reflect the content related to the QA pair in documents. Functionally, question and document keywords work together to pinpoint essential information snippets in the document, with question keywords required to appear in the generated question. However, existing studies overlook the distinction between the two keyword types, which poses a challenge to further improving performance \cite{fei-etal-2022-cqg,DBLP:conf/emnlp/XiaG0YHLN23,ding-etal-2024-sgcm}. 

To break through the above limitation, we define dual-perspective keywords: question keywords, which capture the intent of the questioner, and document keywords, which reflect the content related to the QA pair in documents (Figure \ref{fig01}(a)). Based on this, we further propose a \textbf{D}ual-\textbf{P}erspective \textbf{K}eyword-\textbf{G}uided (DPKG) framework, which seamlessly integrates keywords into the multi-hop question generation process (Figure \ref{fig01}(b)). The framework clarifies essential information snippets by differentiating the two categories of keywords and ensures that specific keywords appear in the generated question, aligning closely with natural human intuition. More specifically, the DPKG framework consists of three main components: (1) an expanded transformer encoder that processes the document and target answer; (2) a decoder that generates dual-perspective keywords; and (3) another decoder that generates multi-hop questions by leveraging crucial information snippets pinpointed by keywords. This framework tackles MQG challenges and aligns with practical requirements. Our contributions are as follows:

\begin{itemize}
\item We define dual-perspective keywords: question keywords, which capture the questioner's intent, and document keywords, which reflect the content related to QA pairs in documents.
\item We propose a DPKG framework that seamlessly integrates keywords into the multi-hop question generation process.
\item Extensive experiments demonstrate the effectiveness of our work, showcasing its promising performance and underscoring its significant value in the MQG task \footnote{https://github.com/imaodong/DPKG}. 
\end{itemize}

\section{Related Work}
Question generation (QG), also known as shallow QG, involves generating questions based on given documents and target answers. Early approaches to QG were rule-based \cite{heilman-smith-2010-good, chali-hasan-2012-towards, mazidi-nielsen-2014-linguistic}, but their performance was limited. As the field advanced, attention-based sequence-to-sequence architectures gained popularity \cite{du-etal-2017-learning,10.1007/978-3-319-73618-1_56, sun-etal-2018-answer,song-etal-2018-leveraging}. Some studies emphasized feature-rich encoders to capture more contextual information \cite{10.1007/978-3-319-73618-1_56,sun-etal-2018-answer}, and answer-aware features were introduced to extract additional information \cite{sun-etal-2018-answer,song-etal-2018-leveraging}. To address the challenge posed by long passages, a maxout pointer mechanism was developed \cite{zhao-etal-2018-paragraph}. Moreover, reinforcement learning (RL) has been applied in a graph-to-sequence model (G2S) for question generation \cite{DBLP:conf/iclr/0022WZ20, chen2023toward}. All of these studies focus on shallow QG, with the SQuAD dataset \cite{rajpurkar-etal-2016-squad} being the most commonly used benchmark.

Multi-hop question generation (MQG) focuses on generating questions that synthesize multiple information snippets from documents to derive target answers. Compared to shallow QG, MQG is more demanding and has garnered significant attention in recent years \cite{su-etal-2020-multi,fei-etal-2021-iterative,fei-etal-2022-cqg,chen2023toward,ding-etal-2024-sgcm}. In the early stages of MQG, graph neural networks (GNNs) were integrated into multi-hop reasoning steps to facilitate question generation \cite{su-etal-2020-multi,pan-etal-2020-semantic}. However, with the evolution of transformer-based architectures, transformers have been demonstrated to outperform GNNs in this task \cite{vaswani2017attention,su2022qa4qg}. To further enhance the performance of pre-trained language models (PLMs) in multi-hop question generation, researchers have incorporated mechanisms such as additional question-answer modules and multi-level content planning \cite{su2022qa4qg,DBLP:conf/emnlp/XiaG0YHLN23}. Numerous studies have also explored auxiliary tasks, proposing methods like order-agnostic learning and sequential rewriting to improve question quality \cite{murakhovska-etal-2022-mixqg,kim-etal-2024-non,DBLP:conf/coling/Hwang0L24}. More recently, question-specific keywords have been utilized during decoding to enhance high-quality generation \cite{fei-etal-2022-cqg}, and \citet{ding-etal-2024-sgcm} supported multi-hop question generation by identifying salient, related keywords or sentences. However, they do not fully harness the guiding potential of keywords, which limits improvements in generation performance.
\section{Methodology}
\begin{figure*}[t]
  \centering
  \includegraphics[width=1.0\textwidth]{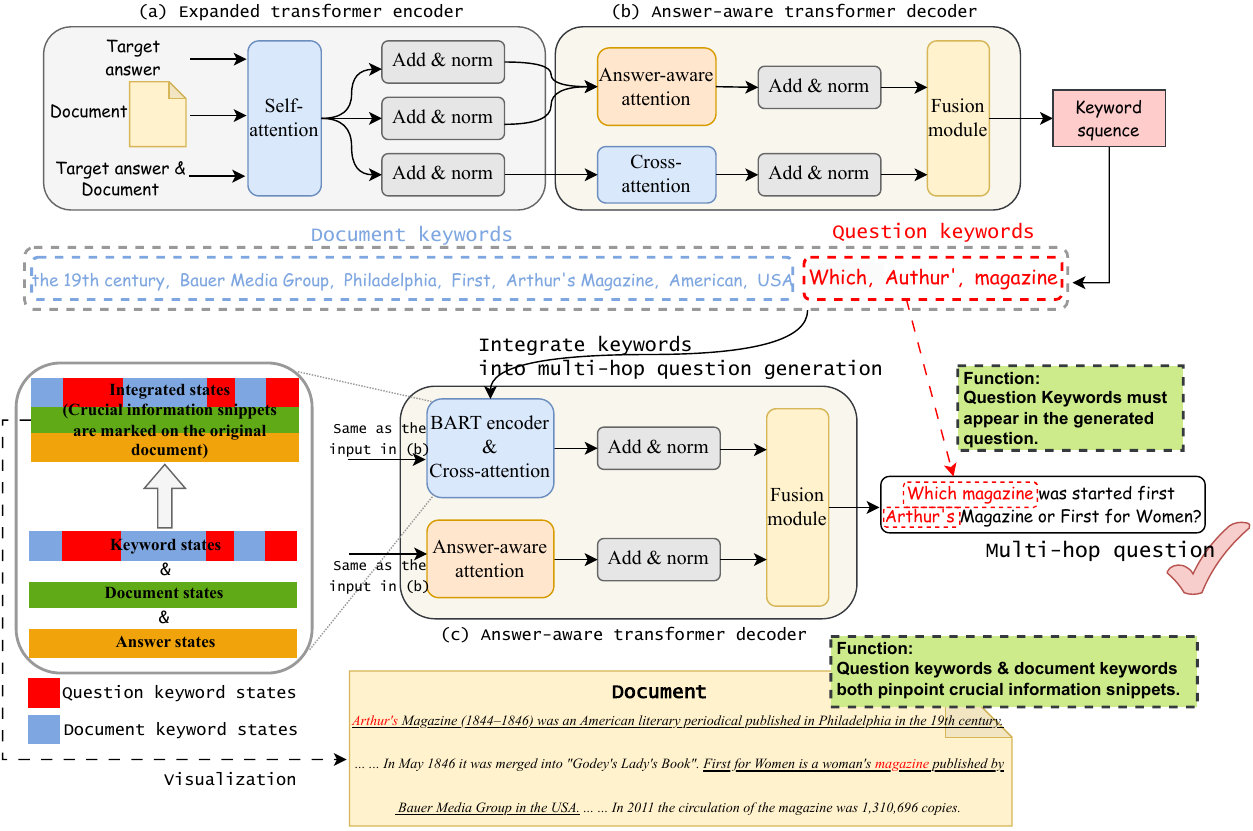}
  \caption{Overall architecture of DPKG framework. In (c), both modules receive the same inputs as in (b), originating from (a). To simplify the illustration and highlight key points, some Transformer details—such as the Feed-Forward sub-layer—are omitted. The encoder includes three Feed-Forward and Add\&Norm sub-layers, and the decoder adds a Fusion module.}
  \label{fig02}
\end{figure*}
\subsection{Task Formulation}
Assume a MQG dataset $U = {\{D, A, Q\}}_{N}^{i=1}$, where $D^i$ denotes contextual documents, $A^i$ represents a target answer, and $N$ is the number of samples. The task is to generate a multi-hop question $Q^i$ based on $D^i$ and $A^i$. We extend the dataset to $U = {\{D, A, Q, K\}}_{N}^{i=1}$, where $K^i$ is a keyword sequence consisting of two sets: $K^q_{j=1:M^q_i}$ and $K^d_{j=1:M^d_i}$, representing question keywords and document keywords annotated with SpaCy (see \S \ref{4.2} for details). Here, $M^q_i$ and $M^d_i$ indicate the number of question keywords and document keywords, respectively. The updated task now includes two primary objectives: (1) generating keywords, $K^i = f_{\theta_1}(D^i, A^i)$; and (2) generating multi-hop question, $Q^i = f_{\theta_2}(D^i, A^i, K^i)$, where $\theta_1$ and $\theta_2$ are model parameters.
\subsection{DPKG Framework}
In this subsection, we introduce the DPKG framework in detail, which seamlessly integrates keywords into the multi-hop question generation process. As illustrated in Figure \ref{fig02}, the DPKG framework consists of an expanded transformer encoder and two answer-aware transformer decoders, designated for keyword generation and question generation, respectively. Additionally, inspired by the success of previous prompt-based approaches, we adopt a similar method for this task. We prepend an additional prefix defined as follows: $D^i$ = "Document: " + $D^i$, $A^i$ = "Answer: " + $A^i$, and $Q^i$ = "Question: " + $Q^i$.

As illustrated in Figure \ref{fig02}(a), this is an expansion of the transformer encoder \cite{vaswani2017attention}, which encodes three components and produces the corresponding hidden states:
\begin{equation}
    H^i_{doc}, H^i_{ans}, H^i_{da} = Encoder_a([D^i, A^i, D^i; A^i])
\end{equation}
where $H^i_{doc}$ and $H^i_{ans}$ represent document states and answer states, respectively. $H^i_{da}$ denotes document-answer states, and $[;]$ denotes concatenation. Notably, we modify the internal structure of the transformer encoder to ensure that, when encoding the three input components simultaneously, each part remains independent, except for the shared self-attention mechanism. We encode the document and answer separately to facilitate the subsequent computation of answer-aware states. In Figure \ref{fig02}, (b) and (c) are identical, except (c) incorporates an additional vanilla BART encoder \cite{lewis-etal-2020-bart} to encode keywords. Although structurally similar, they are independent because of their distinct tasks. Specifically, (b) generates keywords based on the document and answer, while (c) leverages the keywords to pinpoint essential information snippets related to QA pairs in the document and generate multi-hop questions. Formalizing these processes:
\begin{equation}
    K^i = Decoder_{b}(H^i_{doc}, H^i_{ans}, H^i_{da})
\end{equation}
\begin{equation}
    Q^i = Decoder_{c}(H^i_{doc}, H^i_{ans}, H^i_{da}, E(K^i))
\end{equation}
where $E(\cdot)$ denotes the vanilla BART encoder and $K^i$ represents the $i$-th keyword sequence. Additionally, we have modified the transformer decoder by incorporating an answer-aware attention mechanism to enhance answer-awareness in the generation of both keywords and questions.

\subsection{Answer-aware Attention Mechanism and Fusion Module}
\label{AA}
It is worth noting that (b) and (c) in Figure \ref{fig02} are similar processes to each other, with (b) used to generate keywords and (c) used to generate questions. Here, we use (b) as an illustrative example. Assume that in the previous step, we obtain the keyword representation, denoted as $H_{k}^{t-1}$. The answer-aware attention mechanism \cite{10042178} is formalized as follows:
\begin{equation}
    H_{a} = softmax(\frac{H_{k}^{t-1}H_{doc}^T}{\sqrt{d}}\odot K_{weight})H_{doc}
\end{equation}
\begin{equation}
    K_{weight} = MeanPooling(\frac{H_{doc}H_{ans}^T}{\sqrt{d}})
\end{equation}
where $d$ represents the hidden size and $\odot$ denotes element-wise multiplication. We argue that $K_{weight}$ encapsulates the relationship between the document and the answer. By applying the softmax function, we quantify these interconnections. Finally, we take the dot product with $H_{doc}$ and derive the answer-aware states, denoted as $H_{a}$. Correspondingly, in the cross-attention mechanism \cite{vaswani2017attention}, we also obtain hidden states, denoted by $H_{h}$. Next, we apply the fusion module to combine the states of $H_a$ and $H_{h}$, producing the keyword representation $H_{k}^t$ in the following step. The fusion module operates as follows:
\begin{equation}
    H_{k}^{t} = gate \odot H_{a} + (1-gate) \odot H_{h}
\end{equation}
\begin{equation}
    gate = sigmoid([H_{a}; H_{h}])
\end{equation}
\subsection{Keyword Guidance}
\label{3.3}
The keyword guidance concept of DPKG integrates keywords into the multi-hop question generation process by pinpointing crucial information snippets related to QA pairs in the documents and specifying the question keywords that must appear in the generated questions. Our keyword guidance approach is grounded in two primary principles: First, after generating keywords, they are encoded by the BART encoder and aggregated with $H^i_{da}$ to form new states, seamlessly incorporating keywords into the question generation process. This integration enables the identification of crucial information snippets and ensures the inclusion of specific keywords in the generated question, as shown in Figure \ref{fig02}. Second, keyword generation and question generation are jointly trained using a shared, expanded transformer encoder, which allows for deep integration of representations from both tasks.

There are two modes when using dual-perspective keywords in the DPKG framework: (1) In hard mode, special prefixes <qes> or <doc> are added to each keyword to indicate whether it is a question or document keyword. (2) In soft mode, no special prefixes are added; instead, the framework dynamically identifies each keyword's role (question or document) during question generation. This means that, in hard mode, the recognition of the two types of keywords occurs during the keyword generation process, whereas in soft mode, the identification happens during question generation. The soft mode has already achieved impressive performance in other areas of NLP \cite{liu2024coferagcomprehensivefullchainevaluation}. The keywords in this paper are presented sequentially. In hard mode, the keyword sequence is defined as $K^i$=<qes>$K^q_{j=1:M^q_i}$<doc>$K^d_{j=1:M^d_i}$; while in soft mode, it is defined as $K^i$=$K^q_{j=1:M^q_i}K^d_{j=1:M^d_i}$, where the framework dynamically identifies each keyword's role (question or document) during multi-hop question generation.
\subsection{Training Loss}
\label{3.5}
We present a joint training approach for keyword and multi-hop question generation. During training, keyword and question generation are trained in a pipeline fashion, where ground-truth keywords are used for question generation. In the inference phase, keywords are first generated and then used for question generation, creating a gap between training and inference. To bridge this gap, we introduce a novel loss function, defined as follows:
\begin{equation}
    \mathcal{L}_3 = \|F_k-F_g\|_2
\end{equation}
\begin{equation}
    F_k=AvgPooling(H_{k}^t)
\end{equation}
\begin{equation}
    F_g=AvgPooling(E(g))
\end{equation}
where $g$ denotes the ground-truth keywords. This implies that, during joint training, we aim for the ground-truth keyword representations used in multi-hop question generation to closely resemble the generated keyword representations, thereby minimizing this gap. In joint training, the loss for keyword generation is denoted as $\mathcal{L}_1$, while the loss for question generation is denoted as $\mathcal{L}_2$. Both losses are computed using cross-entropy. The final loss is expressed as follows:
\begin{equation}
    \mathcal{L} = \beta_1\mathcal{L}_1 + \beta_2\mathcal{L}_2 + \beta_3\mathcal{L}_3
\end{equation}
where $\beta_1$, $\beta_2$ and $\beta_3$ are hyperparameters \footnote{Experimental details are in Appendix \ref{details}.}. 
\section{Experiments}
\subsection{Dataset}
\label{4.2}
\begin{figure}[t]
  \includegraphics[width=\columnwidth]{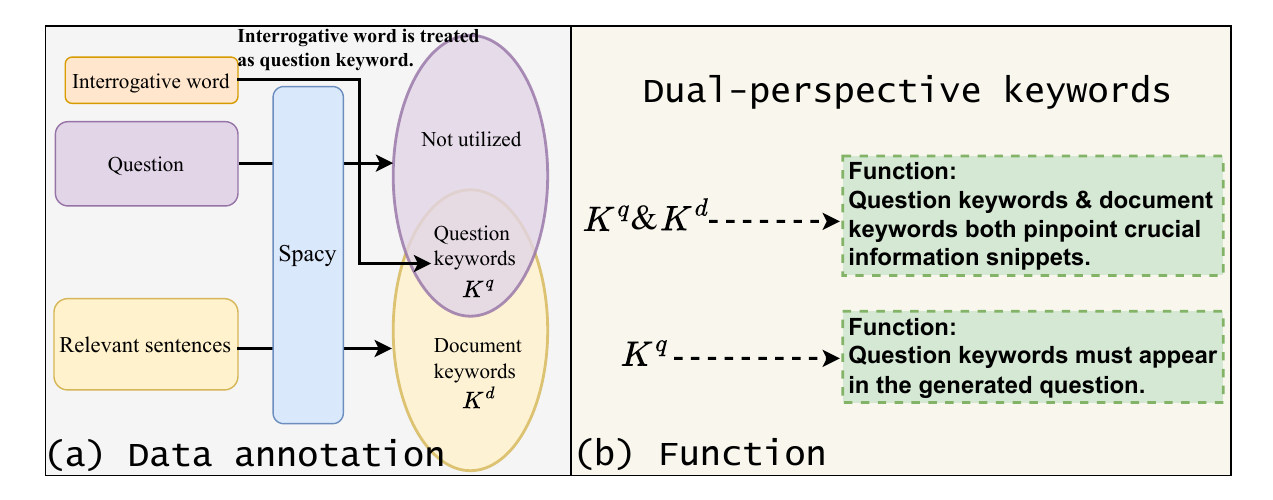}
  \caption{(a) shows our data annotation process; (b) illustrates the functions of question keywords and document keywords. $K^q$ and $K^d$ denote question keywords and document keywords, respectively.}
  \label{figdata}
\end{figure}

We conduct experiments on a common MQG dataset, HotpotQA \cite{yang-etal-2018-hotpotqa}, which consists of two settings: supporting fact (SF) and full document context (Full). In the SF setting, only the sentences containing supporting facts for the answers are provided, whereas in the Full setting, several paragraphs from different documents are presented, which offers a more challenging evaluation of the model's capabilities. Since the original test set is not publicly available, we use the split version provided by \cite{fei-etal-2022-cqg}. We annotate question keywords and document keywords in the HotpotQA dataset using SpaCy \footnote{https://spacy.io/. We utilize the en\_core\_web\_sm model.}. During this process, we focus exclusively on sentences relevant to both answers and questions to ensure that extracted keywords are directly related to them. Entities or phrases, drawn from both relevant sentences and questions, serve as keywords. Figure \ref{figdata}(a) illustrates our process of annotating keywords in detail. Notably, we observe that interrogative words (such as "what," "how," and so on) play a crucial role in the questioning process; therefore, we treat them as question keywords. Through this process, we effectively construct both question and document keywords, ensuring that both types align with the description provided. The details of the keyword statistics are provided in Table \ref{table1}.

However, in Figure \ref{figdata}(a), it is found that a portion is marked as "Not utilized," which means that we do not adopt this portion's keywords. This is because, after questioners view the answer and the document, they will pinpoint the crucial answer-related information snippets in the document while selecting the keywords from those snippets that are relevant to the question (i.e., keywords that must appear in the generated question). Hence, both question and document keywords should be included in the document. This is an intuitive explanation, and empirical explanations are offered in Appendix \ref{appendix:A}.

\begin{table}
\renewcommand{\arraystretch}{0.9}
\small
\centering
\setlength{\tabcolsep}{1.0mm}{
\begin{tabular}{@{}lccccccc@{}}
\toprule
      & \#Sam. & \multicolumn{3}{c}{\#Qes.} & \multicolumn{3}{c}{\#Doc.} \\
      &          & Min.        & Max.        & Avg.       & Min.        & Max.        & Avg.        \\ \midrule
Train & 89947    & 1           & 38          & 6          & 0          & 90          & 9           \\
Dev   & 500      & 1           & 21          & 6          & 1          & 36          & 9           \\
Test  & 7405     & 1           & 20          & 5          & 0          & 55          & 10          \\ \bottomrule

\end{tabular}}
\caption{Statistics of HotpotQA, where "Sam." denotes "Sample," "Qes." denotes "Question keyword," and "Doc." denotes "Document keyword."}
\label{table1}
\end{table}

\subsection{Baselines and Evaluation Metrics}
\begin{table*}[!t]
\renewcommand{\arraystretch}{0.9}
\small
\centering
\setlength{\tabcolsep}{4.0mm}{
\begin{tabular}{lccc|ccc}
\toprule
            & \multicolumn{3}{c|}{Support Fact Sentences (SF)} & \multicolumn{3}{c}{Full Document Context (Full)} \\
Model       & B4             & MTR            & R-L            & B4             & MTR            & R-L            \\ \midrule
BART \cite{lewis-etal-2020-bart}        & 20.39          & 23.46          & 37.23          & 16.77          & 20.07          & 33.69          \\
MulQG \cite{su-etal-2020-multi}       & -              & -              & -              & 15.20          & 20.51          & 35.30          \\
CQG \cite{fei-etal-2022-cqg}         & 25.09          & 27.45          & 41.83          & 21.46              & 24.97              & 39.61              \\
MixQG \cite{murakhovska-etal-2022-mixqg}       & 25.45          & 26.36          & 43.21          & 22.13          & 23.78          & 41.21          \\
QA4QG-large \cite{su2022qa4qg} & 25.70          & 27.44          & \underline{46.48}          & 21.21          & \underline{25.53}          & 42.44          \\
E2EQR \cite{DBLP:conf/coling/Hwang0L24} & 21.73 & 27.47 & 41.34 & - & - & - \\
SGCM \cite{ding-etal-2024-sgcm}   &     26.16          & \textbf{28.51}          & 44.06          & 22.61          & \textbf{26.04}          & 40.61          \\
(Ours) TS-BART &  25.59 & 27.21 & 44.68 & 19.89 & 22.28 & 41.33 \\

\rowcolor[gray]{0.9}
(Ours) DPKG$_{hard}$        & \textbf{26.80}          & \underline{27.87}          & \textbf{46.50}          & \underline{22.74}          & 24.90          & \textbf{43.29}          \\ 
\rowcolor[gray]{0.9}
(Ours) DPKG$_{soft}$     &    \underline{26.19}    & \textbf{28.51}   &  46.36   & \textbf{23.33}   &  25.21   &   \underline{43.18}  \\ \bottomrule
\end{tabular}}
\caption{Performance comparison between DPKG and baselines. \textbf{Bold} text highlights the best results, while \underline{underlined} text indicates the second-best. "B4" denotes "BLEU-4," "MTR" denotes "METEOR," and "R-L" denotes "ROUGE-L."}
\label{table2}
\end{table*}

In our experiments, we compare several advanced models from recent years: \textbf{BART} \cite{lewis-etal-2020-bart}, an encoder-decoder pre-trained model designed for natural language generation; \textbf{MulQG} \cite{su-etal-2020-multi}, which incorporates graph neural networks into multi-hop reasoning to facilitate question generation; \textbf{DP-Graph} \cite{pan-etal-2020-semantic}, a generative model based on Graph2Seq to generate questions; \textbf{CQG} \cite{fei-etal-2022-cqg}, a model that controls key entities during decoding to ensure high-quality generation; \textbf{MixQG} \cite{murakhovska-etal-2022-mixqg}, a generative model trained on diverse question-answering datasets with various answer types; \textbf{QA4QG} \cite{su2022qa4qg}, an enhanced version of BART augmented with a QA module to further constrain generation; \textbf{E2EQR} \cite{DBLP:conf/coling/Hwang0L24}, an end-to-end question rewriting model that increases question complexity through sequential rewriting; \textbf{SGCM} \cite{ding-etal-2024-sgcm}, which supports question generation by identifying salient and relevant sentences. In addition, we introduce \textbf{TS-BART} (Two-stage-BART). In the first stage, BART is fine-tuned to generate dual-perspective keywords; in the second stage, these keywords are integrated into the same BART to generate multi-hop questions. TS-BART is designed to further validate the effectiveness of both the dual-perspective keywords and the overall DPKG framework.


In line with previous studies, we primarily evaluate the DPKG framework using BLEU-4 \cite{DBLP:conf/acl/PapineniRWZ02}, METEOR \cite{banerjee-lavie-2005-meteor}, ROUGE-L \cite{lin-2004-rouge}, and BERTScore \cite{DBLP:conf/iclr/ZhangKWWA20}. 

\subsection{Results and Analysis}
Our comparison between the DPKG framework and advanced MQG models is presented in Table \ref{table2}. In the SF setting, DPKG$_{hard}$ achieves the highest BLEU-4 score of 26.80 and ROUGE-L score of 46.50, while DPKG$_{soft}$ attains the best METEOR score of 28.51. Both DPKG$_{hard}$ and DPKG$_{soft}$ achieve results comparable to current leading levels. In the Full setting, DPKG$_{hard}$ achieves the highest ROUGE-L score of 43.29, while DPKG$_{soft}$ achieves the highest BLEU-4 score of 23.33. DPKG$_{hard}$ and DPKG$_{soft}$ secure the best or second-best performance on BLEU-4 and ROUGE-L, although their performance on METEOR is slightly worse. Across the six evaluation metrics, DPKG$_{hard}$ and DPKG$_{soft}$ rank first or second in most cases, especially in BLEU-4 and ROUGE-L. These results demonstrate that the DPKG framework excels at local matching and capturing long-range semantic dependencies, as reflected in its higher BLEU-4 and ROUGE-L scores.

On the whole, the DPKG framework demonstrates promising performance compared to recent advanced models, achieving state-of-the-art (SOTA) results in most cases. Notably, it outperforms the representative model CQG, which relies on question keywords during the generation process, thereby highlighting the positive impact of dual-perspective keywords in guiding multi-hop question generation. Although TS-BART demonstrates competitive results in the SF setting, its performance in the Full setting is relatively weaker. Nevertheless, it still outperforms BART by a large margin, further demonstrating the advantages of dual-perspective keywords and the overall effectiveness of the DPKG framework. These results underscore the feasibility and value of our dual-perspective keywords and the DPKG framework for the MQG task. It is worth noting that in the SF setting, DPKG$_{hard}$ slightly outperforms DPKG$_{soft}$, while in the Full setting, DPKG$_{soft}$ slightly outperforms DPKG$_{hard}$. This suggests that the role of keyword-guided question generation differs between the two modes, and we will analyze the effects of different keywords on question generation performance in detail in the next subsection.

\begin{table*}[!t]
\centering
\renewcommand{\arraystretch}{0.9}
\small
\setlength{\tabcolsep}{4.0mm}{
\begin{tabular}{@{}lccccccccc@{}}
\toprule
        & \multicolumn{5}{c|}{w/ generated keywords (w/ gk)}               & \multicolumn{4}{c}{w/ ground-truth keywords (w/ gt)} \\ \midrule
        & \multicolumn{9}{c}{Support Fact Sentences (SF)}                                                           \\ \cmidrule(l){2-10} 
Model   & \multicolumn{1}{c|}{KP}    & B4    & MTR   & R-L   & \multicolumn{1}{c|}{BS}    & B4         & MTR       & R-L       & BS        \\ \midrule
DPKG$_{hard}$ & \multicolumn{1}{c|}{\underline{88.13}} & \textbf{26.80} & \underline{27.87} & \textbf{46.50} & \multicolumn{1}{c|}{\textbf{54.14}} & \underline{39.08}      & \underline{35.80}     & \textbf{62.61}     & \textbf{67.34}     \\
DPKG$_{soft}$ & \multicolumn{1}{c|}{\textbf{88.86}} & \underline{26.19} & \textbf{28.51} & \underline{46.36} & \multicolumn{1}{c|}{\underline{54.05}} & \textbf{39.22}      & 34.58     & \underline{61.98}     & \underline{66.51}     \\
DPKG\_D & \multicolumn{1}{c|}{87.05} & 23.95 & 25.20 & 45.62 & \multicolumn{1}{c|}{52.88} & 27.04      & 27.33     & 47.88     & 55.30     \\
DPKG\_Q & \multicolumn{1}{c|}{79.36} & 25.77 & 27.24 & 45.16 & \multicolumn{1}{c|}{53.24} & 38.24      & \textbf{36.41}     & 57.57     & 65.03     \\ \midrule
        & \multicolumn{9}{c}{Full Document   Context (Full)}                                                          \\ \cmidrule(l){2-10} 
DPKG$_{hard}$ & \multicolumn{1}{c|}{\underline{85.39}} & \underline{22.74} & \underline{24.90} & \textbf{43.29} & \multicolumn{1}{c|}{\textbf{50.68}} & \textbf{37.67}      & \textbf{33.28}     & \textbf{60.63}     & \textbf{65.35}     \\
DPKG$_{soft}$ & \multicolumn{1}{c|}{\textbf{86.74}} & \textbf{23.33} & \textbf{25.21} & \underline{43.18} & \multicolumn{1}{c|}{\underline{50.26}} & \underline{35.90}      & 32.06     & \underline{59.68}     & \underline{64.06}     \\
DPKG\_D & \multicolumn{1}{c|}{84.53} & 18.62 & 20.83 & 41.52 & \multicolumn{1}{c|}{48.19} & 20.20      & 21.91     & 42.97     & 49.88     \\
DPKG\_Q & \multicolumn{1}{c|}{77.85} & 21.52 & 23.92 & 42.53 & \multicolumn{1}{c|}{49.90} & 34.24      & \underline{32.64}     & 55.61     & 62.95     \\ \bottomrule
\end{tabular}}
\caption{Effect of different keywords on question generation performance. "BS" denotes "BERTScore."}
\label{table3}
\end{table*}

\begin{figure*}[!t]
  \centering
  \includegraphics[width=1.0\textwidth]{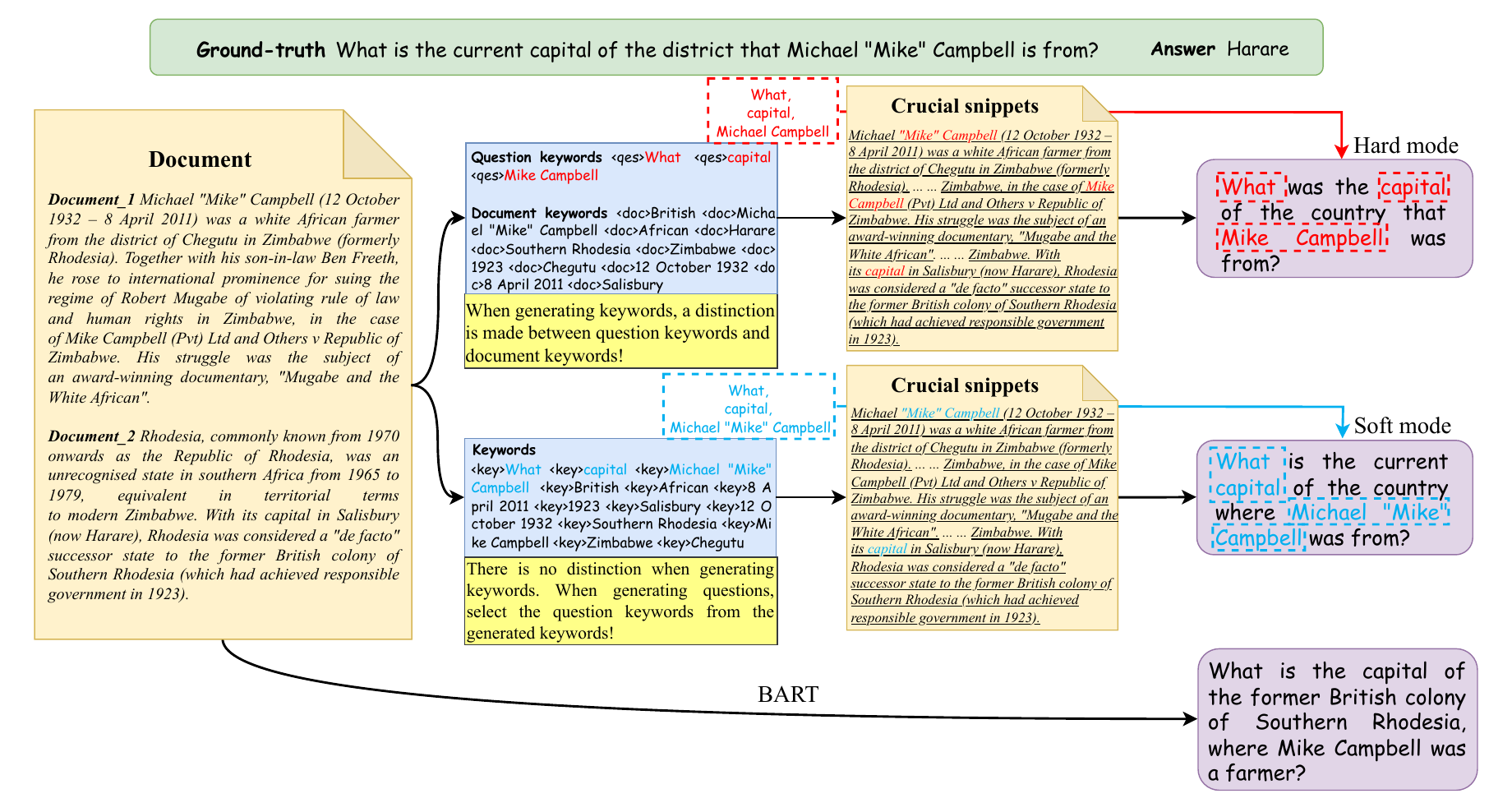}
  \caption{An example of multi-hop question generation is provided, where the \underline{underlined text} highlights crucial information snippets related to the QA pair pinpointed by dual-perspective keywords. The \textcolor{red}{red} denotes the question keywords in hard mode, while the \textcolor{cyan}{blue} represents the question keywords in soft mode.
  }
  \label{fig05}
\end{figure*}

\subsection{Effect of Different Keywords on Results}
\label{4.5}
In this subsection, we examine the effects of different keywords on multi-hop question generation performance. In addition to DPKG$_{hard}$ and DPKG$_{soft}$, we introduce two new model variants: DPKG\_D, which uses only document keywords in the DPKG framework, and DPKG\_Q, which uses only question keywords. Table \ref{table3} shows the performance of these model variants. "KP" in the table denotes keyword generation performance, evaluated using BERTScore. We evaluate DPKG's question generation performance using self-generated keywords ("w/ generated keywords," abbreviated as "w/ gk") and ground-truth keywords ("w/ ground-truth keywords," abbreviated as "w/ gt"). This allows us to assess its ability to generate questions in practice and evaluate the effectiveness of keyword-guided question generation.

As shown in Table \ref{table3}, the question generation performance of DPKG\_Q significantly surpasses that of DPKG\_D in both the SF and Full settings, despite its slightly lower keyword generation performance. Furthermore, when ground-truth keywords are used, DPKG\_Q still outperforms DPKG\_D. This reveals that question keywords tend to capture richer information than document keywords in question generation. The keyword generation performance of DPKG\_D and DPKG\_Q is lower than that of DPKG$_{hard}$ and DPKG$_{soft}$, indicating that considering both types of keywords improves keyword generation performance. While DPKG\_Q has a relatively lower keyword generation performance, its question generation performance remains competitive. However, this does not imply that considering only question keywords is better than considering both, as even with ground-truth keywords, DPKG$_{hard}$ and DPKG$_{soft}$ consistently outperform DPKG\_Q. DPKG$_{hard}$ and DPKG$_{soft}$ rank first or second in nearly all cases, demonstrating that guidance from dual-perspective keywords significantly enhances question generation performance.

With generated keywords, DPKG$_{hard}$ generally outperforms DPKG$_{soft}$ in the SF setting, except for its marginally worse performance on METEOR. Conversely, DPKG$_{soft}$ performs particularly well in the Full setting, especially on BLEU-4 and METEOR. In terms of keyword generation performance, DPKG$_{soft}$ slightly exceeds DPKG$_{hard}$, particularly in the Full setting. This suggests that DPKG$_{soft}$ excels with generated questions mainly due to its superior keyword generation performance. However, DPKG$_{hard}$ still achieves the best performance in the SF setting, despite its slightly lower keyword generation performance. With ground-truth keywords, DPKG$_{hard}$ consistently and significantly outperforms DPKG$_{soft}$. These results highlight that dual-perspective keywords offer more powerful guidance in hard mode than in soft mode.

Moreover, the superior performance of DPKG$_{hard}$ with ground-truth keywords, especially in the Full setting, highlights the ability of dual-perspective keywords to effectively pinpoint crucial information snippets in long documents with abundant noise. Hard mode incorporates special prefixes to indicate each keyword's role during keyword generation, making the process more challenging. This explains why keyword generation performance is slightly worse in hard mode compared to soft mode. In contrast, soft mode requires identifying each keyword's role and generating questions during question generation, which provides weaker guidance than hard mode. These analyses demonstrate the importance of further improving keyword generation performance in hard mode to enhance question generation performance. We also investigate the performance of large language models (LLMs) on multi-hop question generation, as shown in Appendix \ref{promptllm}.

\begin{table}[!t]
\small
\renewcommand{\arraystretch}{0.9}
\centering
\setlength{\tabcolsep}{4.0mm}{
\begin{tabular}{@{}lcccc@{}}
\toprule
        & \multicolumn{4}{c}{Support Fact Sentences (SF)}  \\ \cmidrule(l){2-5} 
Model   & B4          & MTR        & R-L       & BS       \\ \midrule
DPKG$_{hard}$ & \textbf{26.80} & \underline{27.87} & \textbf{46.50} & \textbf{54.14} \\
\hspace{1em}w/o $\mathcal{L}_3$  & 24.74 & 27.83 & 45.08 & 52.93 \\
\hspace{1em}w/o AA  & 24.83 & 27.80 & 45.09 & 52.94 \\
DPKG$_{soft}$ & \underline{26.19} & \textbf{28.51} & \underline{46.36} & \underline{54.05} \\
\hspace{1em}w/o $\mathcal{L}_3$  & 24.59 & 27.56 & 44.93 & 52.68 \\
\hspace{1em}w/o AA  & 24.84 & 27.28 & 44.82 & 52.65 \\ \midrule
        & \multicolumn{4}{c}{Full Document Context (Full)}      \\ \cmidrule(l){2-5} 
DPKG$_{hard}$ & \underline{22.74} & \underline{24.90} & \textbf{43.29} & \textbf{50.68} \\
\hspace{1em}w/o $\mathcal{L}_3$  & 21.16 & 23.90 & 42.01 & 49.61 \\
\hspace{1em}w/o AA  & 21.70 & 24.31 & 42.15 & 49.57 \\
DPKG$_{soft}$ & \textbf{23.33} & \textbf{25.21} & \underline{43.18} & \underline{50.26} \\
\hspace{1em}w/o $\mathcal{L}_3$  & 21.42 & 24.04 & 41.77 & 49.20 \\
\hspace{1em}w/o AA  & 22.41 & 24.89 & 42.59 & 50.05 \\ \bottomrule
\end{tabular}}
\caption{Ablation study.}
\label{table4}
\end{table}

\subsection{Ablation Study}
In this subsection, we conduct an ablation study to verify the effectiveness of each mechanism in the DPKG framework, as shown in Table \ref{table4}. We examine the following mechanisms in the ablation experiments: (1) without the $\mathcal{L}_3$ loss (\textbf{w/o $\mathcal{L}_3$}); and (2) without the answer-aware attention mechanism (\textbf{w/o AA}). From the results in Table \ref{table4}, we observe that both $\mathcal{L}_3$ and the answer-aware attention mechanism significantly contribute to the performance of DPKG$_{hard}$ and DPKG$_{soft}$ in both the SF and Full settings. These findings demonstrate that $\mathcal{L}_3$ effectively bridges the gap discussed in \S \ref{3.5}, while the answer-aware attention mechanism, as detailed in \S \ref{AA}, enhances the quality of the generated questions. The synergistic interaction of DPKG's mechanisms further elevates overall performance.

\subsection{Case Study}
\label{4.7}
To present the quality of multi-hop question generation across different models and to emphasize the role of keywords in guiding this process, we conduct a case study. We focus on DPKG$_{hard}$, DPKG$_{soft}$, and BART, with BART being unguided by keywords. The case is presented in Figure \ref{fig05}. 

It is clear that BART, lacking keyword guidance, generates questions of lower quality. In contrast, both DPKG$_{hard}$ and DPKG$_{soft}$ benefit from keyword guidance and produce higher-quality questions. Specifically, in hard mode, question keywords directly specify terms that should appear in the question, while both question and document keywords work together to pinpoint essential information snippets related to the QA pair in the document, resulting in more concise and clearer questions. In soft mode, DPKG dynamically identifies each keyword's role during question generation, leading to more flexible questions. While both DPKG$_{hard}$ and DPKG$_{soft}$ have their unique characteristics, they pinpoint the same crucial information snippets in the document. Overall, compared with the model without keyword guidance, the DPKG framework generates more understandable and answerable questions, demonstrating the effectiveness of both the DPKG framework and dual-perspective keywords. We also conduct extended analytical experiments; see Appendix \ref{more_case_analysis} for details.

\section{Conclusion}
In this paper, we define dual-perspective keywords: question keywords, which capture the questioner's intent, and document keywords, which reflect the content related to QA pairs in documents. Furthermore, we propose a DPKG framework that seamlessly integrates keywords into the multi-hop question generation process. We analyze the role of different keywords in guiding question generation and demonstrate that dual-perspective keywords are effective, especially in hard mode, providing more powerful guidance. Extensive experiments demonstrate the effectiveness of our work, showcasing its promising performance and underscoring its significant value in the MQG task.
\section*{Limitations}
We categorize keywords into question keywords and document keywords and propose a DPKG framework to seamlessly integrate them into the multi-hop question generation process. Although extensive experiments have demonstrated the effectiveness of our work, several limitations remain, such as the limited performance of the proposed keywords in guiding large language models (LLMs) for multi-hop question generation (Appendix \ref{promptllm}). This is because multi-hop question generation requires not only a better understanding of semantics but also alignment with the answers. LLMs are often prone to hallucinations, and recent studies have shown that training dedicated PLMs can achieve better results than using large models \cite{ushio-etal-2023-practical,liu2024syntheticcontextgenerationquestion}. Eliminating hallucinations when prompting LLMs to generate multi-hop questions will be an interesting challenge. Additionally, the focus of our study is on MQG, also known as deep question generation (DQG) \cite{pan-etal-2020-semantic, fei-etal-2022-cqg, DBLP:conf/emnlp/XiaG0YHLN23}. Suitable datasets for this research are limited, highlighting the need to develop new, appropriate datasets, which is highly meaningful.
\bibliography{custom}

\newpage
\appendix
\section{Experimental Details}
\label{details}
We employ AdamW as the optimizer, incorporating warmup and gradient clipping techniques. The batch size is set to 8, with a learning rate of $1 \times 10^{-5}$. Training is conducted for 50 epochs with an early stopping strategy. The hyperparameters $\beta_1$, $\beta_2$, and $\beta_3$ are all set to 1.0, and the beam size is 8. All experiments are performed on a GeForce RTX 3090 GPU.

In our DPKG framework, the encoder and decoder are modified based on the transformer architecture, with most of their parameters being compatible with BART’s pre-trained model. Therefore, during training, we load the parameters compatible with BART’s pre-trained model \footnote{https://huggingface.co/facebook/bart-base}, initialize the incompatible parameters with a mean of 0 and a variance of 0.02, and then fine-tune all parameters. Additionally, it is worth noting that the hyperparameter settings not discussed in this paper are identical to those of BART-base, including the number of encoder and decoder layers, among others.
\section{Exploring Different Data Annotation Methods}
\label{appendix:A}
\begin{table}[!t]
\small
\centering
\setlength{\tabcolsep}{1.0mm}{
\begin{tabular}{@{}lccccccc@{}}
\toprule
      & \#Sam. & \multicolumn{3}{c}{\#Qes.} & \multicolumn{3}{c}{\#Doc.} \\
      &          & Min.        & Max.        & Avg.       & Min.        & Max.        & Avg.        \\ \midrule
Train & 89947    & 1           & 52          & 9          & 0          & 96          & 13           \\
Dev   & 500      & 2           & 45          & 8          & 2          & 43          & 13           \\
Test  & 7405     & 1           & 24          & 8          & 0          & 63          & 14          \\ \bottomrule
\end{tabular}}
\caption{Statistics of HotpotQA (version 2), where "Sam." denotes "Sample," "Qes." denotes "Question keyword," and "Doc." denotes "Document keyword."}
\label{appendix_t1}
\end{table}

\begin{table*}[!t]
\centering
\small
\setlength{\tabcolsep}{4.0mm}{
\begin{tabular}{@{}lccccccccc@{}}
\toprule
        & \multicolumn{5}{c|}{w/   generated keywords (w/ gk)}                             & \multicolumn{4}{c}{w/   ground-truth keywords (w/ gt)} \\ \midrule
        & \multicolumn{9}{c}{Support   Fact Sentences (SF)}                                                                                        \\ \cmidrule(l){2-10} 
Model   & \multicolumn{1}{c|}{KP}    & B4    & MTR   & R-L   & \multicolumn{1}{c|}{BS}    & B4           & MTR         & R-L         & BS          \\ \midrule
DPKG$_{hard}$ & \multicolumn{1}{c|}{\underline{88.13}} & \textbf{26.80} & \underline{27.87} & \textbf{46.50} & \multicolumn{1}{c|}{\textbf{54.14}} & 39.08        & \underline{35.80}       & \underline{62.61}       & \underline{67.34}       \\
DPKG$_{soft}$ & \multicolumn{1}{c|}{\textbf{88.86}} & \underline{26.19} & \textbf{28.51} & \underline{46.36} & \multicolumn{1}{c|}{\underline{54.05}} & \underline{39.22}        & 34.58       & 61.98       & 66.51       \\
DPKG$_{v2}$    & \multicolumn{1}{c|}{86.38} & 25.79 & 26.95 & 45.94 & \multicolumn{1}{c|}{53.34} & \textbf{52.34}        & \textbf{42.73}       & \textbf{76.09}       & \textbf{78.31}       \\ \midrule
        & \multicolumn{9}{c}{Full   Document Context (Full)}                                                                                       \\ \cmidrule(l){2-10} 
DPKG$_{hard}$ & \multicolumn{1}{c|}{\underline{85.39}} & \underline{22.74} & \underline{24.90} & \textbf{43.29} & \multicolumn{1}{c|}{\textbf{50.68}} & \underline{37.67}        & \underline{33.28}       & \underline{60.63}       & \underline{65.35}       \\
DPKG$_{soft}$ & \multicolumn{1}{c|}{\textbf{86.74}} & \textbf{23.33} & \textbf{25.21} & \underline{43.18} & \multicolumn{1}{c|}{\underline{50.26}} & 35.90        & 32.06       & 59.68       & 64.06       \\
DPKG$_{v2}$    & \multicolumn{1}{c|}{83.48} & 18.40 & 21.05 & 41.53 & \multicolumn{1}{c|}{47.98} & \textbf{51.77}        & \textbf{42.45}       & \textbf{75.44}       & \textbf{77.73}       \\ \bottomrule
\end{tabular}}
\caption{Performance comparison. \textbf{Bold} text highlights the best results, while \underline{underlined} text indicates the second-best. "B4" denotes "BLEU-4," "MTR" denotes "METEOR," "R-L" denotes "ROUGE-L," "BS" denotes "BERTScore" and "KP" stands for "keyword generation performance."}
\label{appendix_t2}
\end{table*}
In this paper, we use the method shown in Figure \ref{figdata}(a) to annotate question keywords and document keywords. However, a question arises: why is there a "Not utilized" portion? While we provide an intuitive explanation in the main text, this subsection offers an empirical explanation.

We consider the "Not utilized" portion in Figure \ref{figdata}(a) as the question keywords. In this way, question keywords are extracted from the question, and document keywords are extracted from the relevant sentences. Question keywords and document keywords are independent, with no overlap. Additionally, interrogative words (such as "what," "how," and so on) are treated as question keywords. The statistics for version 2 annotations are shown in Table \ref{appendix_t1}. We observe that the number of question keywords and document keywords in this version has increased significantly compared to those in Table \ref{table1}. We use the annotated dataset here and adopt the hard mode of DPKG, with the new model called DPKG$_{v2}$. The performance comparison with DPKG$_{hard}$ and DPKG$_{soft}$ is shown in Table \ref{appendix_t2}.

From the results in Table \ref{appendix_t2}, it is worth noting that the multi-hop question generation performance of DPKG$_{v2}$ with generated keywords is significantly worse than that of DPKG$_{hard}$ and DPKG$_{soft}$, especially in the Full setting. The performance of keyword generation remains poor. These results indicate that increasing the number of keywords hinders the performance of keyword generation and further inhibits the improvement of question generation performance. DPKG$_{v2}$ performs significantly better under ground-truth keywords, outperforming both DPKG$_{hard}$ and DPKG$_{soft}$. This is intuitive, as both the question and document keywords in this annotation method provide more detailed information. For example, when generating the question, all terms related to it are included (with the "Not utilized" portion treated as question keywords). In summary, while this annotation method provides more information, its practical performance is often unsatisfactory. The data annotation method used in the main text strikes a trade-off: it achieves SOTA performance in practice while also providing valuable information. This empirically explains why the "Not utilized" portion is included in the annotated approach of our main text.


\section{Results of Large Language Models on Question Generation}
\label{promptllm}
\subsection{Results of Prompting Large Language Models}
\begin{table}[!t]
\small
\centering
\setlength{\tabcolsep}{3.0mm}{
\begin{tabular}{@{}lcccc@{}}
\toprule
          & \multicolumn{4}{c}{Support Fact Sentences (SF)}  \\ \cmidrule(l){2-5} 
Model     & B4          & MTR        & R-L       & BS        \\ \midrule
          & \multicolumn{4}{c}{w/o keyword guidance}          \\ \cmidrule(l){2-5} 
Qwen-plus &             &            &           &           \\
\hspace{2em}w/ zero-shot & 13.60       & 19.56      & 33.88     & 41.92     \\
\hspace{2em}w/ one-shot  & 14.97       & 20.68      & 34.42     & 42.87     \\ \cmidrule(l){2-5} 
          & \multicolumn{4}{c}{w/ keyword guidance}         \\ \cmidrule(l){2-5} 
Qwen-plus &             &            &           &           \\
\hspace{2em}w/ zero-shot & 20.35       & 25.44      & 44.28     & 52.07     \\
\hspace{2em}w/ one-shot  & 22.81       & 27.57      & 46.46     & 54.47     \\
DPKG$_{hard}$   & \underline{39.08}       & \textbf{35.80}      & \textbf{62.61}     & \textbf{67.34}     \\
DPKG$_{soft}$   & \textbf{39.22}       & \underline{34.58}      & \underline{61.98}     & \underline{66.51}     \\ \midrule
          & \multicolumn{4}{c}{Full Document Context (Full)} \\ \cmidrule(l){2-5} 
          & \multicolumn{4}{c}{w/o keyword guidance}          \\ \cmidrule(l){2-5} 
Qwen-plus &             &            &           &           \\
\hspace{2em}w/ zero-shot & 10.62       & 16.82      & 30.34     & 38.11     \\
\hspace{2em}w/ one-shot  & 11.17       & 17.37      & 30.39     & 38.18     \\ \cmidrule(l){2-5} 
          & \multicolumn{4}{c}{w/ keyword guidance}         \\ \cmidrule(l){2-5} 
Qwen-plus &             &            &           &           \\
\hspace{2em}w/ zero-shot & 18.77       & 24.14      & 42.71     & 50.52     \\
\hspace{2em}w/ one-shot  & 20.91       & 26.18      & 44.88     & 52.70     \\
DPKG$_{hard}$   & \textbf{37.67}       & \textbf{33.28}      & \textbf{60.63}     & \textbf{65.35}     \\
DPKG$_{soft}$   & \underline{35.90}       & \underline{32.06}      & \underline{59.68}     & \underline{64.06}     \\ \bottomrule
\end{tabular}}
\caption{Performance comparison between the DPKG framework and prompting LLMs on multi-hop question generation.}
\label{llmprompt_t}
\end{table}

\begin{figure*}[!t]
  \centering
  \includegraphics[width=1.0\textwidth]{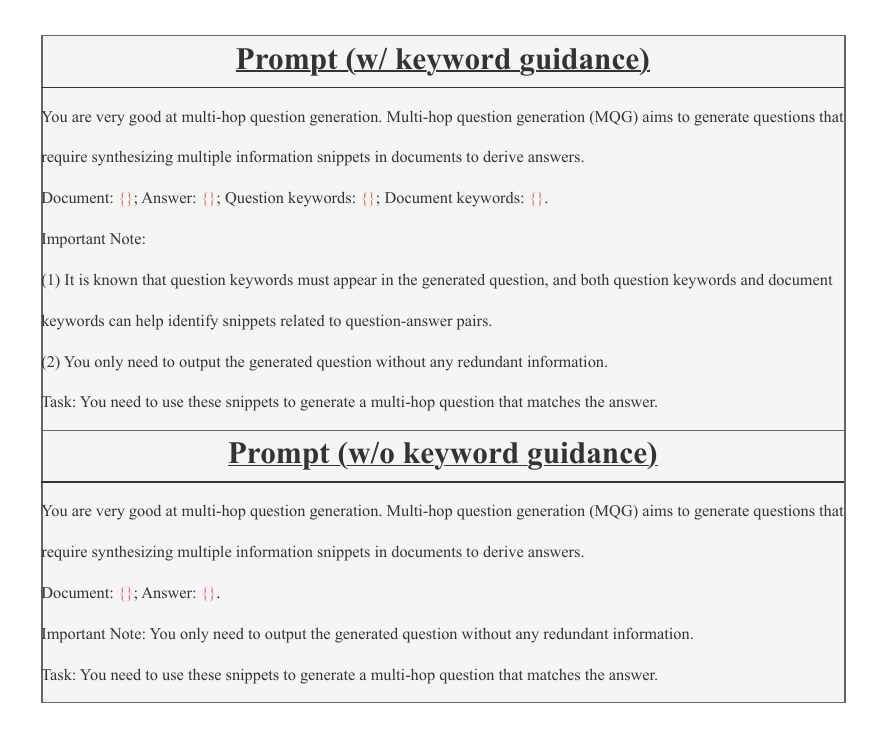}
  \caption{Prompt template.}
  \label{promptfig}
\end{figure*}
In this subsection, we examine the performance of large language models (LLMs) prompted by keywords on multi-hop question generation. Specifically, to explore the guiding role of dual-perspective keywords, we conduct experiments using ground-truth keywords. Our primary focus is the Qwen-Plus model \footnote{https://bailian.console.aliyun.com}, which performs well in English and understands prompt instructions. We evaluate its question generation capabilities in both zero-shot and one-shot settings, using the prompt template illustrated in Figure \ref{promptfig}. The experimental results are presented in Table \ref{llmprompt_t}, where "w/o keyword guidance" means no keyword guidance, and "w/ keyword guidance" refers to dual-perspective keyword guidance.

According to the results in Table \ref{llmprompt_t}, we find that under the guidance of dual-perspective keywords, the quality of LLM-generated questions has greatly improved, and the performance of Qwen-Plus in the one-shot setting has significantly outperformed that in the zero-shot setting. These results show that keyword guidance improves question generation performance and demonstrate the effectiveness of the dual-perspective keywords proposed in this paper. However, the performance of Qwen-Plus in all cases still lags behind that of the DPKG framework proposed in this paper. This is because multi-hop question generation requires not only a better understanding of semantics but also alignment with the answers. LLMs are often prone to hallucinations. Although hallucinations are somewhat mitigated under the guidance of keywords, they remain prevalent. Some studies have shown that training dedicated PLMs can achieve better results than prompting large models \cite{ushio-etal-2023-practical,liu2024syntheticcontextgenerationquestion}. Overall, we find that the dual-perspective keywords presented in this paper effectively prompt LLMs, though they are less effective than the dedicated DPKG framework. These results highlight the urgent need for a framework that effectively prompts LLMs using keywords and eliminates hallucinations when generating multi-hop questions. 

\subsection{Results of Fine-tuned Large Language Models}
\begin{table}[!t]
\small
\centering
\setlength{\tabcolsep}{2.4mm}{
\begin{tabular}{@{}lcccc@{}}
\toprule
                    & B4     & MTR    & R-L   & BS   \\ \cmidrule(l){2-5} 
Model               & \multicolumn{4}{c}{Support Fact Sentences (SF)}  \\ \midrule
Llama-3.1-8B (Lora) & 25.29      & 26.68     & \textbf{46.73}     & \textbf{54.63}       \\
DPKG$_{hard}$          & \textbf{26.80}      & \underline{27.87}     & \underline{46.50}     & \underline{54.14}       \\
DPKG$_{soft}$          & \underline{26.19}      & \textbf{28.51}     & 46.36     & 54.05       \\ \midrule
                    & \multicolumn{4}{c}{Full Document Context (Full)} \\ \cmidrule(l){2-5} 
Llama-3.1-8B (Lora) & 22.49      & 24.67     & \textbf{44.38}     & \textbf{51.98}       \\
DPKG$_{hard}$          & \underline{22.74}      & \underline{24.90}     & \underline{43.29}     & \underline{50.68}       \\
DPKG$_{soft}$          & \textbf{23.33}      & \textbf{25.21}     & 43.18     & 50.26       \\ \bottomrule
\end{tabular}}
\caption{Performance comparison between the DPKG framework and fine-tuned LLMs on multi-hop question generation.}
\label{fine-tunedllm}
\end{table}
In this subsection, we evaluate the performance of fine-tuned LLMs for multi-hop question generation. Specifically, we use Llama-3.1-8B \footnote{https://huggingface.co/meta-llama/Llama-3.1-8B-Instruct} (abbreviated as Llama), fine-tuned via the LoRA method. The results of the experiment are displayed in Table \ref{fine-tunedllm}.

It can be seen from Table \ref{fine-tunedllm} that in both the SF and Full settings, Llama's performance on BLEU-4 and METEOR is inferior to that of the DPKG framework, while it performs better on ROUGE-L and BERTScore. Although Llama has a large number of parameters, it does not demonstrate stronger performance compared to the DPKG framework. On the one hand, this highlights the difficulty of the MQG task; on the other, it underscores the effectiveness of the dual-perspective keywords and the DPKG framework. These findings again suggest the challenges large models face in generating multi-hop questions — a problem that remains to be addressed in future research.


\section{Extended Analysis}
\label{more_case_analysis}
\subsection{Additional Case Analysis}
\begin{table*}[!t]
\begin{tabular}{|l|p{13cm}|} 
\hline
Document & \uline{\textcolor{purple}{Dana Brunetti} (\textcolor{red}{born} June 11, 1973) is an \textcolor{red}{American film producer} and social networking entrepreneur. He is the \textcolor{purple}{president} of \textcolor{purple}{Kevin Spacey's} \textcolor{red}{production} company, \textcolor{red}{Trigger Street Productions}.} In 2016, he became the presidents of Relativity Media with Kevin Spacey taking on a chairman position. \newline \uline{\textcolor{purple}{Kevin Spacey Fowler}, \textcolor{purple}{KBE} (\textcolor{red}{born} \textcolor{purple}{July 26, 1959}) is an \textcolor{red}{American actor}, \textcolor{red}{film director}, \textcolor{red}{producer}, screenwriter, and singer. He began his career as a stage \textcolor{red}{actor} during the 1980s before obtaining supporting roles in \textcolor{red}{film} and television.} He gained critical acclaim in the early 1990s that culminated in his first Academy Award for Best Supporting Actor for the neo-noir crime thriller "The Usual Suspects" (1995), and an Academy Award for Best Actor for midlife crisis-themed drama \uline{"\textcolor{red}{American} Beauty" (1999).} \\ \hline
DPKG$_{hard}$ & \textcolor{red}{When} was the \textcolor{red}{American actor}, \textcolor{red}{film director}, \textcolor{red}{producer}, screenwriter, and singer \textcolor{red}{born} who's \textcolor{red}{production} company is \textcolor{red}{Trigger Street Productions}? \\ \hline
Document & \uline{\textcolor{cyan}{Dana Brunetti} (\textcolor{cyan}{born} June 11, 1973) is an \textcolor{cyan}{American film producer} and social networking entrepreneur. He is the \textcolor{cyan}{president} of \textcolor{purple}{Kevin Spacey's} \textcolor{cyan}{production} company, \textcolor{purple}{Trigger Street Productions}.} In 2016, he became the presidents of Relativity Media with Kevin Spacey taking on a chairman position. \newline \uline{\textcolor{purple}{Kevin Spacey Fowler}, \textcolor{purple}{KBE} (\textcolor{cyan}{born} \textcolor{purple}{July 26, 1959}) is an \textcolor{cyan}{American actor}, \textcolor{cyan}{film director}, \textcolor{cyan}{producer}, screenwriter, and singer. He began his career as a stage \textcolor{cyan}{actor} during the 1980s before obtaining supporting roles in \textcolor{cyan}{film} and television.} He gained critical acclaim in the early 1990s that culminated in his first Academy Award for Best Supporting Actor for the neo-noir crime thriller "The Usual Suspects" (1995), and an Academy Award for Best Actor for midlife crisis-themed drama \uline{"American Beauty" (1999).} \\ \hline
DPKG$_{soft}$ & \textcolor{cyan}{Dana Brunetti} is the \textcolor{cyan}{president} of a \textcolor{cyan}{production} company owned by an \textcolor{cyan}{American actor}, \textcolor{cyan}{film director}, \textcolor{cyan}{producer}, screenwriter, and singer, \textcolor{cyan}{born} on what date? \\ \hline
BART & Dana Brunetti is the president of the production company of an American actor born on what date? \\ \hline
Ground-truth & When was the American actor, film director which Dana Brunetti is the president of his company born \\ \hline
\end{tabular}
\caption{Another case: The \textcolor{red}{red} represents the question keywords in hard mode, while the \textcolor{cyan}{blue} represents those in soft mode. The \textcolor{purple}{purple} indicates the document keywords, and the \underline{underlined text} highlights essential information snippets.}
\label{more_case_table}
\end{table*}

We provide a more detailed analysis here. We have previously examined the specific case of keyword-guided multi-hop question generation (see section \ref{4.7}). In this section, we present another concrete example.

As you can see from Table \ref{more_case_table}, DPKG$_{hard}$ and DPKG$_{soft}$ pinpoint the same essential information snippets. While they identify different question keywords, the resulting questions are multi-hop and contain a wealth of information that can be understood and answered. The table contains rich information, and we use different colors to highlight it. We observe that the same words may be assigned different roles in different modes, which affects the resulting question. For example, "Trigger Street Productions" is identified as a question keyword in hard mode and as a document keyword in soft mode. There are many more cases like this, which demonstrate that hard mode and soft mode have their own characteristics, as we discussed in section \ref{4.7}. In addition, we find that the questions generated by BART are more straightforward, contain less multi-hop information, and are not as rich as DPKG$_{hard}$ and DPKG$_{soft}$. This is mainly because BART does not have keyword guidance, does not know what is essential, and can simply generate questions through the mappings it learns during training.

In short, through the analysis of the above cases, we gain a more intuitive understanding of the necessity of keyword-guided multi-hop question generation, especially the two types of keywords proposed in this paper. The two modes of the framework proposed here each have their own characteristics, revealing their significant value in the MQG task.

\subsection{Results of Generated Questions on QA}
\begin{table}[!t]
\small
\centering
\setlength{\tabcolsep}{3.5mm}{
\begin{tabular}{@{}lcc|cc@{}}
\toprule
              & \multicolumn{2}{c|}{SF} & \multicolumn{2}{c}{Full} \\ \cmidrule(l){2-5} 
Model         & EM                   & F1                  & EM                  & F1                 \\ \midrule
BART          & 54.71                & 70.43               & 54.86               & 69.28              \\
DPKG$_{hard}$    & \textbf{58.76}                & \textbf{73.65}               & \underline{56.31}               & \textbf{71.05}              \\
DPKG$_{soft}$    & \underline{57.19}                & \underline{72.21}               & \textbf{56.79}               & \underline{70.99}              \\ \midrule
Ground-truth & 63.24                & 79.02               & 61.63               & 77.29              \\ \bottomrule
\end{tabular}}
\caption{Results of generated questions on the QA system. "Ground-truth" indicates the use of gold-standard questions.}
\label{QA-hotpot}
\end{table}
To further verify the validity of the generated questions, we evaluate them using a question-answering (QA) system. Specifically, we use Qwen-plus \footnote{https://bailian.console.aliyun.com} as our QA system, to which we input the generated questions and obtain answers. The evaluation metrics we adopt are Exact Match (EM) and F1 score (F1). The results of the experiment are shown in Table \ref{QA-hotpot}. From the experimental results, we can see that even Ground-truth does not achieve strong performance, which may be related to the limitations of the QA system itself. Overall, the DPKG framework achieves competitive results, significantly outperforming BART, though still lagging behind the Ground-truth. These results further validate the effectiveness of the DPKG framework from the perspective of question answering.

\subsection{Human Evaluation}
While automatic evaluation offers valuable insights into model performance, n-gram-based metrics have inherent limitations. Therefore, we conduct a human evaluation by randomly selecting 50 questions from the generated set in the Full setting. Based on experimental results, we evaluate the following representative models: (1) \textbf{DPKG$_{hard}$}; (2) \textbf{DPKG$_{soft}$}; (3) \textbf{TS-BART}; and (4) \textbf{BART}. Three crowdsourced workers rate each sample on a scale from 1 (poor) to 5 (good) across three key dimensions: (i) complexity, which assesses whether the question requires reasoning across multiple snippets in the document; (ii) relevance, which determines if the question is answerable and contextually appropriate; and (iii) fluency, which evaluates whether the question adheres to grammatical rules and maintains coherent logic. We summarize and average the workers' ratings, presenting the results in Table \ref{table-human}.

\begin{table}
\small
\centering
\setlength{\tabcolsep}{0.8mm}{
\begin{tabular}{@{}lcccc@{}}
\toprule
Model        & Complexity & Relevance & Fluency & Average\\ \midrule
BART         & 3.17    & 3.28     & 3.24    & 3.23\\
TS-BART           & 3.68    & 3.75     & 3.62    & 3.68\\
DPKG$_{hard}$      & \textbf{4.54}    & \underline{4.55}     & \underline{4.50}    & \underline{4.53}\\
DPKG$_{soft}$         & \underline{4.53}    & \textbf{4.57}     & \textbf{4.56}    & \textbf{4.55} \\ \midrule
Ground-truth & 4.94    & 4.97     & 4.92 & 4.94    \\ \bottomrule
\end{tabular}}
\caption{Human evaluation results. "Average" denotes the average of complexity, relevance, and fluency. "Ground-truth" indicates the use of gold-standard questions.}
\label{table-human}
\end{table}
According to the results displayed in Table \ref{table-human}, the DPKG framework has achieved very promising results compared with BART and TS-BART. Specifically, BART is a vanilla model that does not use dual-perspective keywords, while TS-BART first uses vanilla BART to generate the keywords, and then reuses it to generate questions guided by those keywords. TS-BART outperforms BART in both automatic and human evaluations, fully demonstrating the guiding role of dual-perspective keywords. The DPKG framework outperforms TS-BART, further confirming its overall effectiveness. DPKG$_{hard}$ performs slightly worse than DPKG$_{soft}$. DPKG$_{hard}$ and DPKG$_{soft}$ perform comparably in terms of Complexity and Relevance, while DPKG$_{soft}$ shows slightly better performance in Fluency. These results indicate that the questions generated by DPKG$_{soft}$ are more fluent in the Full setting, which is consistent with the automatic evaluation results. All in all, the human evaluation results further demonstrate the effectiveness of the dual-perspective keywords and the DPKG framework.

\section{Generalization to Other Datasets}
To further verify the effectiveness of the DPKG framework, we also use another common dataset, MusiQue-2hop \cite{trivedi-etal-2022-musique}, and follow the same keyword annotation procedure for questions and documents as in HotpotQA. Notably, since the test set of MusiQue-2hop is not publicly available, we construct a custom split. Table \ref{musiquedata} provides detailed keyword statistics for MusiQue-2hop. We conduct two types of experiments: comparing our results with current advanced models, and verifying the generated questions using a QA system. For the comparative evaluation, our baselines include—in addition to CQG, MulQG, and E2EQR—DP-Graph \cite{pan-etal-2020-semantic}, a generative Graph2Seq-based question generation model. For the QA evaluation, we use Qwen-plus as our QA system, into which we input the generated questions to obtain answers. Table \ref{musiquemain} shows the results of the comparative evaluation, while Table \ref{musique-QA} presents the QA evaluation results.

\begin{table}[!t]
\small
\centering
\setlength{\tabcolsep}{1.0mm}{
\begin{tabular}{@{}lccccccc@{}}
\toprule
      & \#Sam. & \multicolumn{3}{c}{\#Qes.} & \multicolumn{3}{c}{\#Doc.} \\
      &        & Min.    & Max.    & Avg.   & Min.    & Max.    & Avg.   \\ \midrule
Train & 13124  & 1       & 12      & 3      & 3       & 170     & 24     \\
Dev   & 1252   & 1       & 11      & 3      & 4       & 134     & 24     \\
Test  & 1252   & 1       & 9       & 3      & 4       & 134     & 24     \\ \bottomrule
\end{tabular}}
\caption{Statistics of MusiQue-2hop, where "Sam." denotes "Sample," "Qes." denotes "Question keyword," and "Doc." denotes "Document keyword."}
\label{musiquedata}
\end{table}

\begin{table}[!t]
\small
\centering
\setlength{\tabcolsep}{2.0mm}{
\begin{tabular}{lccc}
\toprule
Model           & B4   & MTR   & R-L   \\ \midrule
DP-Graph \cite{pan-etal-2020-semantic}   & 5.14  & 10.80 & 28.69 \\
CQG \cite{fei-etal-2022-cqg}        & 9.64  & 16.20 & 33.98 \\
MulQG \cite{su-etal-2020-multi}     & 9.56  & 15.41 & 37.18 \\
E2EQR \cite{DBLP:conf/coling/Hwang0L24}      & 20.33 & 25.64 & 44.01 \\
\rowcolor[gray]{0.9}
(Ours) DPKG$_{hard}$ & \underline{28.33} & \underline{27.77} & \underline{51.93} \\
\rowcolor[gray]{0.9}
(Ours) DPKG$_{soft}$ & \textbf{29.03} & \textbf{28.14} & \textbf{52.38} \\ \bottomrule
\end{tabular}}
\caption{Performance comparison between DPKG and baselines on MusiQue-2hop.}
\label{musiquemain}
\end{table}

\begin{table}[!t]
\small
\centering
\setlength{\tabcolsep}{4.0mm}{
\begin{tabular}{@{}lcc@{}}
\toprule
Model         & EM    & F1    \\ \midrule
DPKG$_{hard}$    & \underline{41.45} & \textbf{50.95} \\
DPKG$_{soft}$    & \textbf{41.53} & \underline{50.93} \\ \midrule
Ground-truth & 58.79 & 71.42 \\ \bottomrule
\end{tabular}}
\caption{Results of generated questions evaluated by the QA system on MusiQue-2hop. "Ground-truth" indicates the use of gold-standard questions.}
\label{musique-QA}
\end{table}

Table \ref{musiquemain} shows that the DPKG framework achieves the best performance compared to other models, indicating its effectiveness. More specifically, DPKG$_{soft}$ performs slightly better than DPKG$_{hard}$, consistent with our earlier conclusion. Table \ref{musique-QA} shows that Ground-truth does not perform strongly, which may be due to the limitations of the QA system. Although DPKG$_{soft}$ performs slightly better than DPKG$_{hard}$ on n-gram-based metrics, DPKG$_{hard}$ slightly outperforms DPKG$_{soft}$ on the QA system. This suggests that the QA system itself has some level of semantic understanding. This also highlights the limitations of n-gram-based metrics and the necessity of using the QA system to evaluate the generated questions. To sum up, the experiments on the MusiQue-2hop further demonstrate the effectiveness and generalizability of the DPKG framework.

\end{document}